\documentclass[%
 apl
jmp,%
 amsmath,amssymb,
reprint,%
]{revtex4-1}

\usepackage{natbib}
\usepackage{mathtools}
\usepackage{graphicx}
\usepackage{epstopdf}
\usepackage{enumitem}

\usepackage[caption=false]{subfig}

\usepackage{dcolumn}
\usepackage{bm}

\begin{document}


\title[
]{Magnetoresistive RAM for error resilient XNOR-Nets}

\author{M. Tzoufras}
 \email{mtzoufras@physics.ucla.edu}
 \affiliation{Spin Memory, Inc., Fremont, California 94538, USA}
\author{M. Gajek}
 \affiliation{Spin Memory, Inc., Fremont, California 94538, USA}
\author{A. Walker}
 \affiliation{Spin Memory, Inc., Fremont, California 94538, USA}

\date{\today}

\begin{abstract} 
We trained three Binarized Convolutional Neural Network architectures (LeNet-4, Network-In-Network, AlexNet) on a variety of datasets (MNIST, CIFAR-10, CIFAR-100, extended SVHN, ImageNet) using error-prone activations and tested them without errors to study the resilience of the training process. With the exception of the AlexNet when trained on the ImageNet dataset, we found that Bit Error Rates of a few percent during training do not degrade the test accuracy. Furthermore, by training the AlexNet on progressively smaller subsets of ImageNet classes, we observed increasing tolerance to activation errors. The ability to operate with high BERs is critical for reducing power consumption in existing hardware and for facilitating emerging memory technologies. We discuss how operating at moderate BER can enable Magnetoresistive RAM with higher endurance, speed and density.
\end{abstract}


\maketitle

\section{Introduction}

Artificial Neural Networks (ANNs) are biology-inspired concepts that have in recent years revolutionized many areas of research and industry and even much of everyday life. Managing their power consumption has been one of the key challenges that has accompanied their emergence and especially the advent of Deep Neural Networks (DNNs). When considering the analogy with biological intelligence we find that biology needs $4$ to $5$ orders of magnitude less power, primarily due to to its synaptic operation energy efficiency, at the ``expense" of nearly $75\%$ synaptic error rate \cite{DOEneuromorphic}. In this paper we explore how the presence of errors during training can impact the classification accuracy and discuss how operating at moderate Bit Error Rate (BER) facilitates Magnetoresistive RAM (MRAM) technology for ANN applications.

In all-perpendicular Spin Transfer Torque MRAM (STT-MRAM) \cite{Mangin:2006zv,Ikeda:2010wb}, a bit is stored in a Magnetic Tunnel Junction (MTJ) comprising two ferromagnetic layers separated by a thin insulating barrier. The magnetization vectors of the two ferromagnets are perpendicular to the plane of the layers and may be in a parallel (P) or antiparallel (AP) configuration. When electrical current passes though one of the ferromagnets it gets spin-filtered and the spin-polarized electrons impart spin torque \cite{SLONCZEWSKI1996L1,PhysRevB.54.9353} on the other. In order to read a bit of information we must supply enough current to identify whether the MTJ is in the P (low-resistance) or AP (high-resistance) state but not so high that the  spin torque disturbs the magnetization of either of the layers. Writing a bit requires higher current than reading because one must produce enough spin torque to flip the magnetization of one of the ferromagnets; yet too high a voltage across the MTJ stresses the insulator material and degrades its endurance. The switching process is inherently stochastic and the switching probability can be calculated analytically given the MTJ parameters and the read/write pulse amplitude and duration \cite{Tzoufras:2017bx}. MRAM exhibits many advantages compared to conventional memories, including non-volatility, high endurance and high density, but having to contend with its stochasticity remains a major obstacle to widespread adoption. Therefore, architectures and applications that are resilient to errors are the best candidates for MRAM.

Approximate computing \cite{Venkataramani:2014jb,Murmann:2015yb,Moons:2016xd} has been proposed as a way to trade classification accuracy for energy efficiency in inference tasks. The accuracy-power trade-off was first studied in silicon by \citet{Yang:2017px}, using SRAM with reduced voltage supply to train and test a three-layer Convolutional Neural Net (ConvNet) on the low-complexity MNIST handwritten digit dataset. The presence of BER due to sub-threshold voltage during training produced an increase in the classification accuracy when the SRAM was operated similarly for testing.  In  Ref. \cite{Yang:2018fh}, it was shown that a deeper ConvNet trained on a moderate-complexity dataset, the CIFAR-10, is also resilient to bit errors during inference, albeit less than the three-layer ConvNet trained on MNIST. In 2018, a framework was developed to study DNN resilience during inference \cite{Reagen:2018:AFQ:3195970.3195997} and potential sources of errors were identified in SRAM, DRAM and flash memory. 

Apart from hardware errors, the common practice of limiting the number representation and employing fixed-point arithmetic in neural network applications introduces quantization noise. This approach reduces both memory and compute requirements and has been studied extensively since the 1990s  \cite{Iwata:1989cy,Hammerstrom:1990qw,Holi:1993cr}. Recently, \citet{Gupta:2015:DLL:3045118.3045303} demonstrated that stochastic rounding yields superior performance when using low-precision fixed-point computations compared to the standard round-to-nearest method. Stochastic rounding is also seen as the preferred approach for the extreme case of binary representation that has been garnering increasing interest for inference applications. In Refs. \cite{DBLP:journals/corr/CourbariauxB16,NIPS2016_6573,DBLP:journals/corr/RastegariORF16}, several training algorithms were developed that enable Binarized Neural Networks (BNNs) to achieve---along with drastic reduction in power consumption---classification accuracy comparable to non-binarized networks. Moreover, binarization of the convolution in ConvNets turns it into an XNOR operation which leads to further enhancement in speed and energy efficiency. Accordingly XNOR-Nets are excellent candidates for edge applications, where density and power are most constrained.

Stochastic rounding for a BNN takes the form shown in Ref.  \cite{DBLP:journals/corr/CourbariauxB16}: 
\begin{equation}
x^b = 
\biggl\{
\begin{array}{l}
-1 \quad\mbox{with probability}\quad 1-\sigma(x)
\\
+1 \quad\mbox{with probability}\quad \sigma(x)
\end{array}
\end{equation}
where $\sigma$ is the ``hard sigmoid" function:
$\sigma(x) = \max(0,\min(1,\frac{x+1}{2}))
$, a linear function that performs stochastic rounding in the same manner as suggested in Ref. \cite{Gupta:2015:DLL:3045118.3045303}. 

However, generating the plethora of random numbers needed for stochastic rounding is not practical in most systems and round-to-nearest is usually chosen. Due to its stochastic nature, an MTJ can be used as an alternative tunable true random number generator for stochastic rounding but this also introduces unwieldy complexity in the circuit, namely a digital-to-analog converter to provide the current that corresponds to the desired switching probability. Instead, we examine what happens when the MRAM is operated at a constant reduced voltage, i.e. at fixed BER. This involves no additional complexity compared to standard MRAM. We may write the rounding function due to the MTJ stochasticity as:
\begin{eqnarray}
x^{AP} &=& 
\biggl\{
\begin{array}{l}
-1 \quad\mbox{with probability} \quad p_{-1}, 
\\
+1 \quad\mbox{with probability}\quad 1-p_{-1}
\end{array}
\\
x^{P} &=& 
\biggl\{
\begin{array}{l}
-1 \quad\mbox{with probability} \quad 1-p_{+1}, 
\\
+1 \quad\mbox{with probability}\quad p_{+1}
\end{array}
\end{eqnarray}
where $p_{-1}$  and $p_{+1}$ indicate successful write of the AP and P states respectively. Below we assume for simplicity that $p_{-1}=p_{+1}=p$. 

Aside from their importance for edge applications, XNOR-Nets are suitable for isolating the effect of BER in neural networks because one does not need to worry about protecting the most significant bits or exploring various fault mitigation strategies \cite{Reagen:2016cr}. For the XNOR-Nets studied in this article, when a bit error occurs we ignore it and make no attempt at detecting, mitigating or correcting it. 

Network weights and activations are known to have different tolerance to errors and the same holds true for the individual network layers and training epochs. Specifically, weights are expected to be less resilient to BER than activations such that the effect of weight errors would dominate the outcome if the same BER was present across all of the network variables. Here we only allow for bit errors in the binary activations during training with constant BER across all binary layers and epochs. Our guiding principle is to study the effect of BER in ANNs in the most transparent conditions. Future work will explore the effect of BER on weights as well as the combined effect of weight and activation errors.


\section{Training XNOR-Nets with BER in the binary activations}\label{section:training}

We conducted experiments on three binarized ConvNets and several datasets of increasing complexity: namely the binarized LeNet-4 on the MNIST dataset (section \ref{subsection:LeNet4}), the binarized Network-in-Network on the CIFAR-10, the CIFAR-100 and the extended SVHN datasets (section \ref{subsection:NiN}), and finally the binarized AlexNet trained on the ImageNet dataset as well as several ImageNet subsets (section \ref{subsection:AlexNet}).

\subsection{LeNet-4} \label{subsection:LeNet4}

We first present the classic combination of a LeNet Convolutional Neural Net architecture  \cite{Lecun:1998hs}, one of the simplest ConvNets, training on the low-complexity MNIST dataset. We binarized a modified LeNet-4 architecture comprising: (I) a regular convolution layer, with batch normalization and ReLu activation followed by a max-pooling layer, (II) a binary convolution block that comprises batch normalization, binary activation, and binary convolution followed by a max-pooling layer, (III) a binary fully connected layer, and (IV) a softmax classification layer. This network was trained on the 60,000 train images of the MNIST handwritten digit dataset and tested on its 10,000 test images. During training the filter weights were left error-free while the binary activations exhibited a fixed BER. We examined BERs between $0\%$ and $16\%$ and repeated the training process $10$ times for each value of BER.

\begin{figure}[thbp]
\includegraphics[width=\columnwidth]{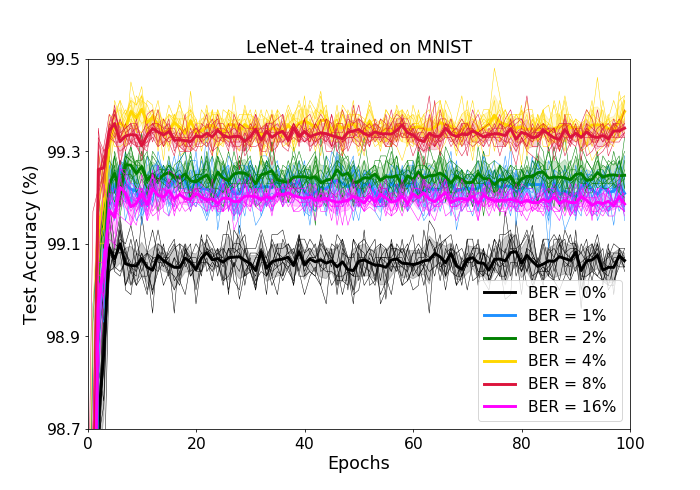}
\caption{\label{fig:MNIST} 
Test accuracy vs training epochs for a binarized LeNet-4 network trained on the MNIST dataset.} 
\end{figure}

The test accuracy is shown in Figure \ref{fig:MNIST} for all of the above experiments and the average for each BER value as well as the individual traces are displayed to give a sense of the spread between consecutive runs. The accuracy gradually improved when raising the BER from $\mbox{BER}=0\%$ (no errors) to $\mbox{BER} = 4\%$, and plateaued between  $\mbox{BER} = 4\%$ and $\mbox{BER} = 8\%$. Increasing the BER beyond this point showed a reduction in the test accuracy. Interestingly, at $BER = 16\%$ the test accuracy was still higher than in the case where no errors were included during training, highlighting the robustness of the training process to the presence of activation errors. In Ref. \cite{Yang:2017px} it was found that matching error rate distributions between training and testing can improve classification accuracy. In contrast, here we find that even without errors during testing the classification accuracy is higher than in the error-free case when the $BER\leq16\%$. (We note however that in Ref. \cite{Yang:2017px} the entire memory, SRAM, was operated at low voltage, while we only studied activation errors.)

\subsection{Network-In-Network}\label{subsection:NiN}

To validate these findings in a more elaborate architecture combined with datasets of higher complexity, we studied the effect of activation errors in the binarized Network-In-Network (NiN)  \cite{DBLP:journals/corr/LinCY13}, a classic architecture that inspired the Inception Networks \cite{Szegedy:2015ol},  using the CIFAR-10, CIFAR-100 and extended Street View House Numbers (SVHN) datasets. The binarized NiN comprises three stages, each stage having three convolution layers followed by a pooling layer (max-pooling, average-pooling, average-pooling respectively for each stage). All convolution layers were binarized except the first and last ones, where ReLu activations were used. The activations of the binarized layers were subject to BER.

\begin{figure}[htp]

\subfloat[Network-In-Network trained on CIFAR-10]{\label{subfig:CIFAR10}
 \includegraphics[width=\columnwidth]{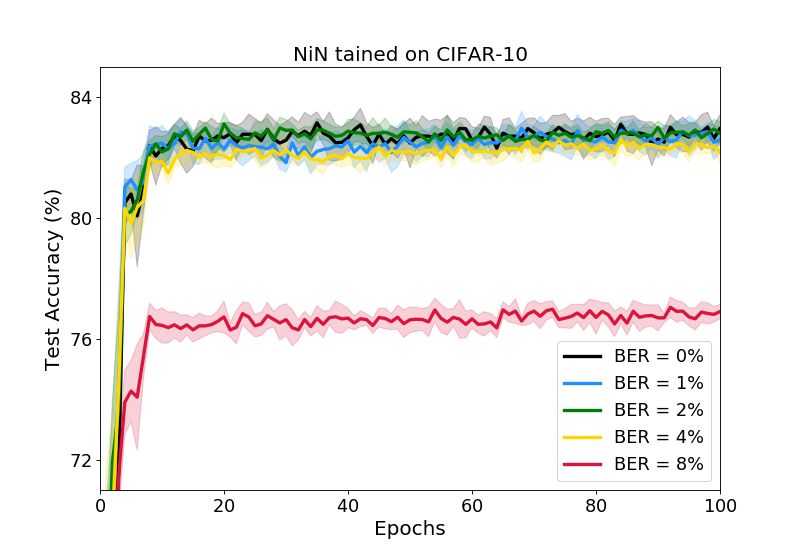}
}

\subfloat[Network-In-Network trained on CIFAR-100]{\label{subfig:CIFAR100}
 \includegraphics[width=\columnwidth]{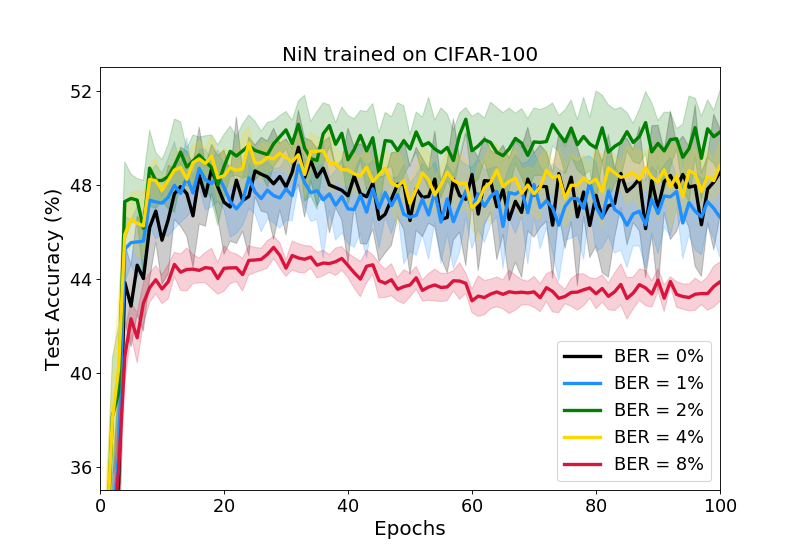}
 }
 
\subfloat[Network-In-Network trained on the extended SVHN]{\label{subfig:SVHN}
  \includegraphics[width=\columnwidth]{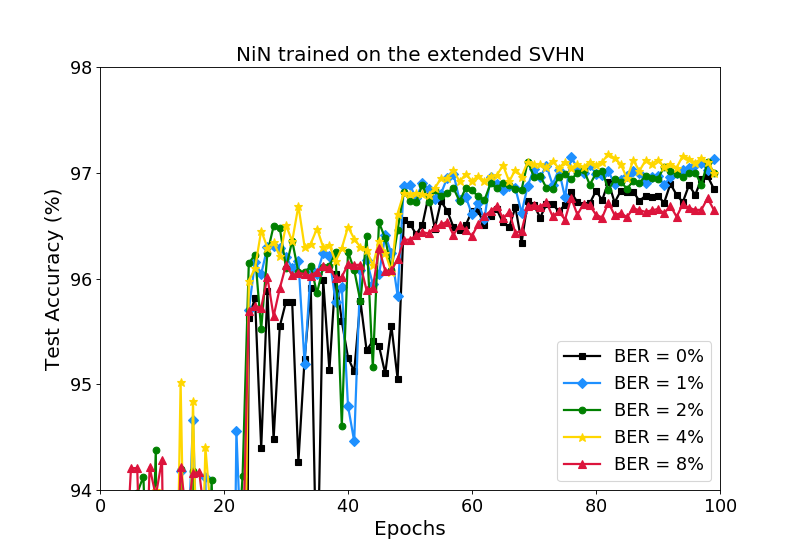}
}

\caption{ \label{fig:CIFARs}Test accuracy vs training epochs for a binarized Network-In-Network trained on (a) the CIFAR-10, (b) the CIFAR-100, and (c) the extended SVHN datasets. For the latter dataset we present the raw data from the experiment.}
\end{figure}

The CIFAR-10 dataset is of moderate complexity and contains 50,000 train and 10,000 test images in RGB with size $32\times32$ that belong to $10$ classes. 
Figure \ref{subfig:CIFAR10} shows the test accuracy when using BER between $0\%-8\%$ for the binary activations. The results plotted are averages over $10$ experiments and the standard deviation is also included. We find that for low BER values, up to $\mbox{BER}= 2\%$, there is no noticeable degradation in test accuracy but at $\mbox{BER}=4\%$ there is a visible drop and at $8\%$ the drop is very significant.

The CIFAR-100 dataset comprises 100 classes with 500 train and 100 test images per class in RGB with size $32\times32$. Due to the higher number of classes and smaller number of examples per class compared to the CIFAR-10 dataset we achieved lower test accuracy when training the binarized NiN on the CIFAR-100. The results (averages over 10 experiments) are shown in Figure \ref{subfig:CIFAR100}.  
Similarly to the two previous cases we observe an initial rise of  the test accuracy combined with a drop below the maximum for $BER= 4\%$. Additionally we note that optimal performance was reached for $BER=2\%$ and that for higher BERs, e.g. $BER=8\%$, the standard deviation was visibly reduced.

The same NiN architecture was trained on the extended SVHN dataset, which contains 531,131 train and 26,032 test images, size $32\times32$, RGB,  belonging to $10$ classes, one for each digit. This is a more complex dataset than MNIST and it contains a much larger number of train images. The findings of this experiment are akin to the previous experiments and displayed in Figure \ref{subfig:SVHN}. We find a slight improvement in test accuracy with increasing BER up to $4\%$ followed by a drop when further raising the BER.

\subsection{AlexNet}\label{subsection:AlexNet}

We now turn to the ImageNet Large-Scale Visual Recognition Challenge which contains a train set of more than 1.2M images and a test set of 60,000. This dataset includes 1000 categories of about 1000 images each, with size $224\times224$.  We trained a binarized AlexNet architecture \cite{Krizhevsky:2017:ICD:3098997.3065386,DBLP:journals/corr/RastegariORF16} which incorporates 5 convolutional layers, the first of which is the only one that is not binarized. Max-pooling layers are used after the first, second and fifth convolutional layers. This implementation achieved a Top1 classification accuracy of $44.07\%$, virtually identical to the one reported in Ref. \cite{DBLP:journals/corr/RastegariORF16}.

\begin{figure}[htp]

\subfloat[Binarized AlexNet trained on the ImageNet dataset]{\label{subfig:ILSVRC2012}
 \includegraphics[width=\columnwidth]{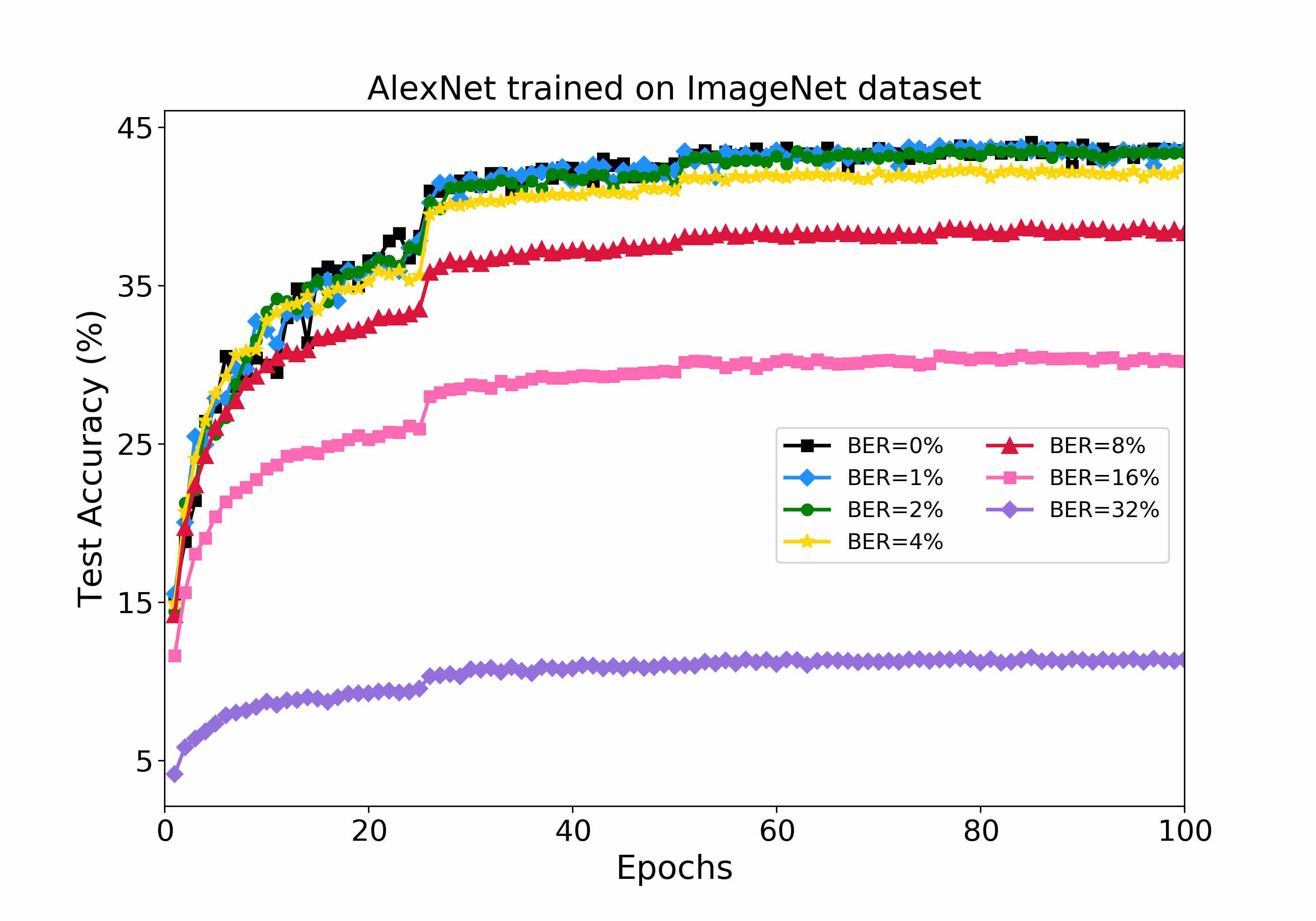}
}

\subfloat[Binarized AlexNet trained on a 100-class subset of ImageNet]{\label{subfig:ImageNet100}
 \includegraphics[width=\columnwidth]{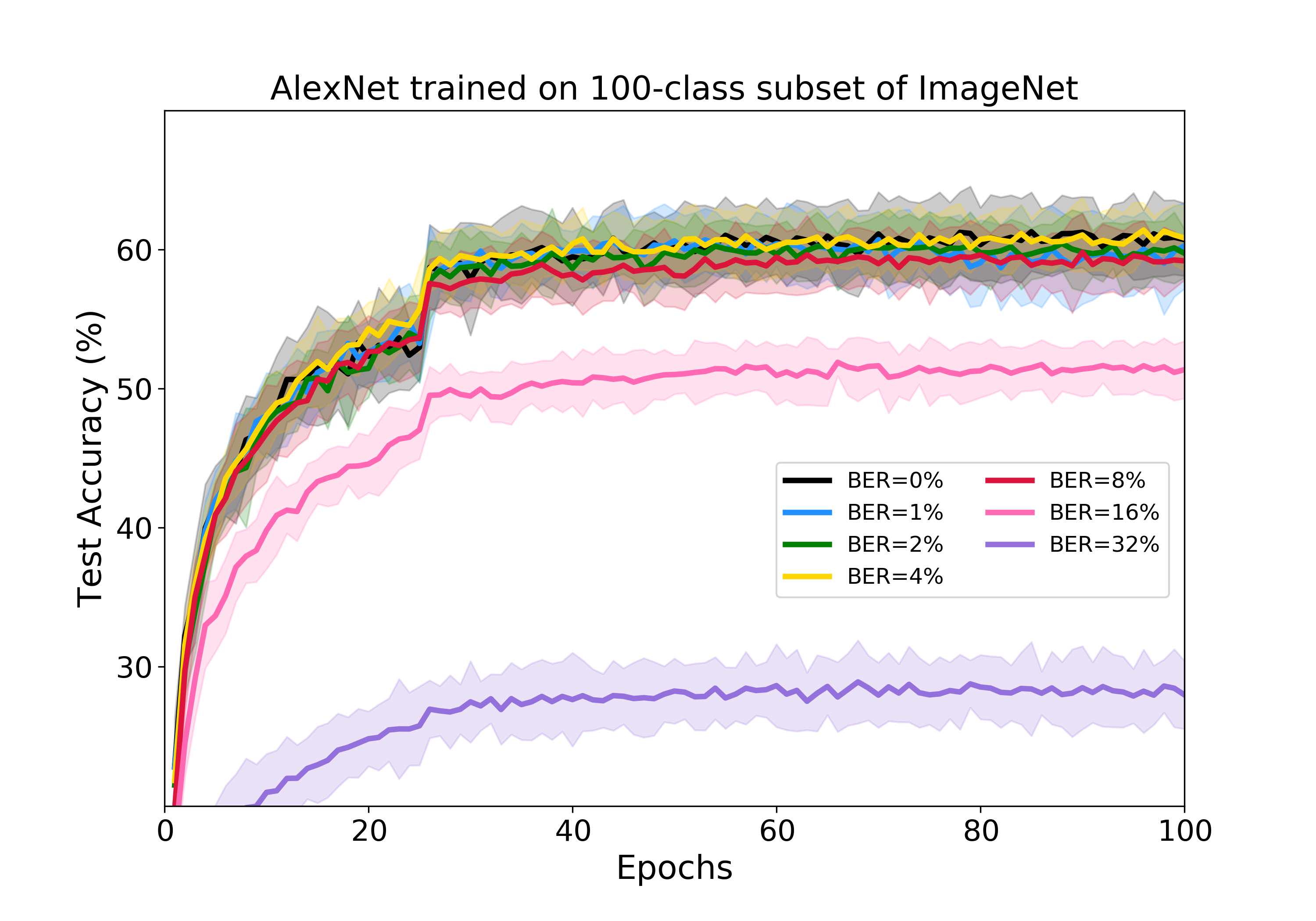}
 }
 
\subfloat[Binarized AlexNet trained on a 10-class subset of ImageNet]{\label{subfig:ImageNet10}
  \includegraphics[width=\columnwidth]{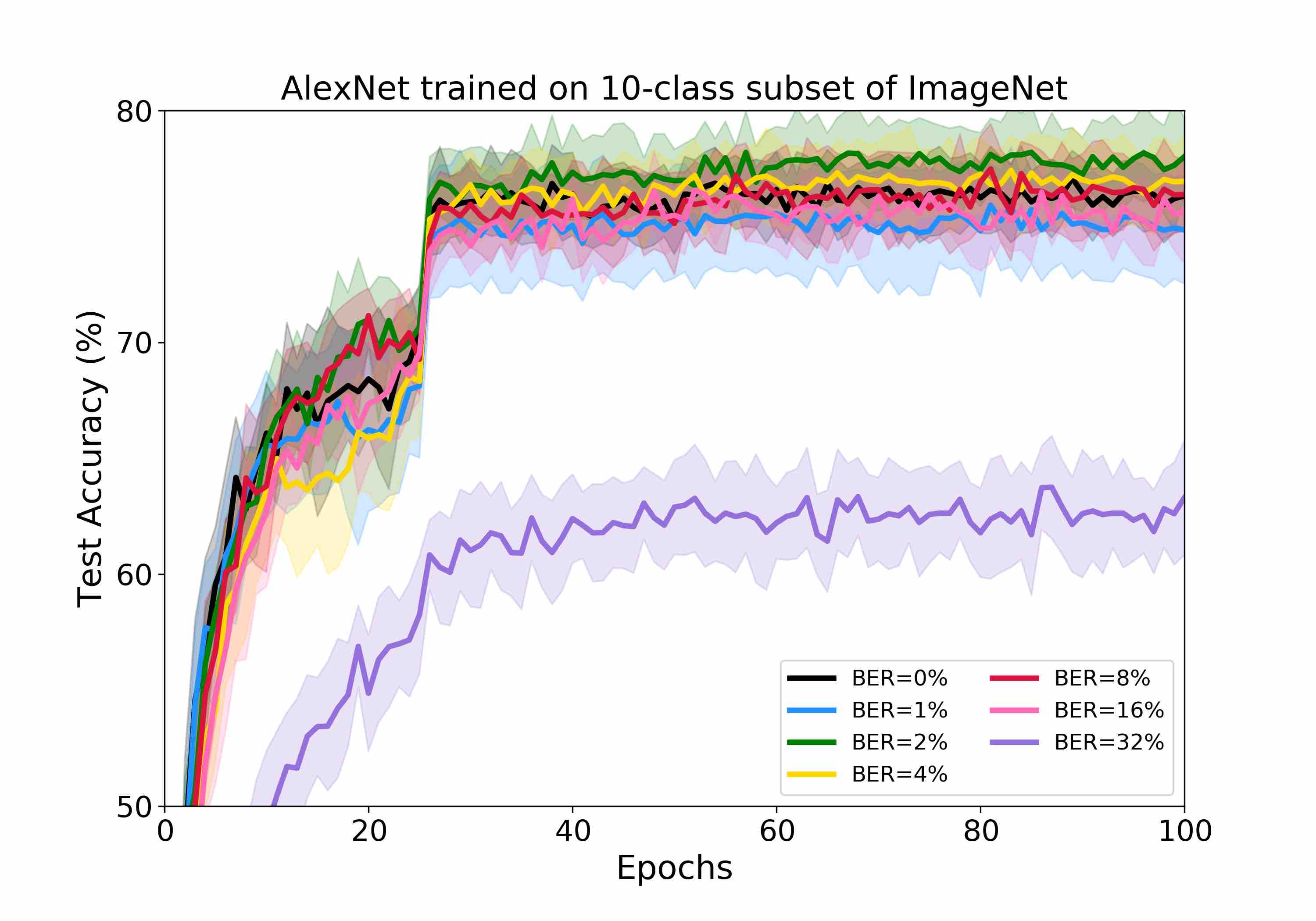}
}

\caption{ \label{fig:AlexNet}Test accuracy vs training epochs for a binarized AlexNet architecture trained on (a) the 1000-class ImageNet dataset including BER of $0\%-32\%$ 
in the binary activations, (b) a randomly selected 100-class subset of the ImageNet and (c) a 10-class subset of the ImageNet. 
For (b) and (c) we ran 10 experiments and show the average and the typical dispersion. }
\end{figure}

In contrast to our experiments in sections \ref{subsection:LeNet4}-\ref{subsection:NiN}  there is no discernible increase in test accuracy when raising the binary activation BER up to $2\%$ during training. At $\mbox{BER} = 4\%$ there was a noticeable drop and beyond $4\%$ the performance continues degrading rapidly. Results of this training process are shown in Figure \ref{subfig:ILSVRC2012}.

In order to isolate the influence of the network architecture from the complexity of the dataset we selected a random 100-class subset of the 1000-class dataset and repeated the training for various BER values. Each experiment was run 10 times and the average along with the typical dispersion are shown in Figure \ref{subfig:ImageNet100}. The test accuracy exceeded $60\%$ because there were fewer classes and therefore fewer semantic neighbors. Unlike the 1000-class dataset, the 100-class subset showed no significant decline in test accuracy up to $\mbox{BER} = 8\%$. A second 100-class subset was randomly chosen (not shown) and the experiment qualitatively replicated the behavior seen in Figure \ref{subfig:ImageNet100} from the first 100-class subset.

In a subsequent experiment we used a randomly-selected 10-class subset of ImageNet, further increasing the semantic distance between classes. In Fig  \ref{subfig:ImageNet10}, the training process shows enhanced resiliency to BER compared to the 100-class subset. No degradation in accuracy was seen up to $\mbox{BER} = 16\%$. A second experiment (not shown) using a separate randomly-selected 10-class subset of ImageNet replicated this behavior. 

\begin{figure}[htp]

\subfloat[Binarized AlexNet trained on a 2-class subset of ImageNet]{\label{subfig:ImageNet2}
 \includegraphics[width=\columnwidth]{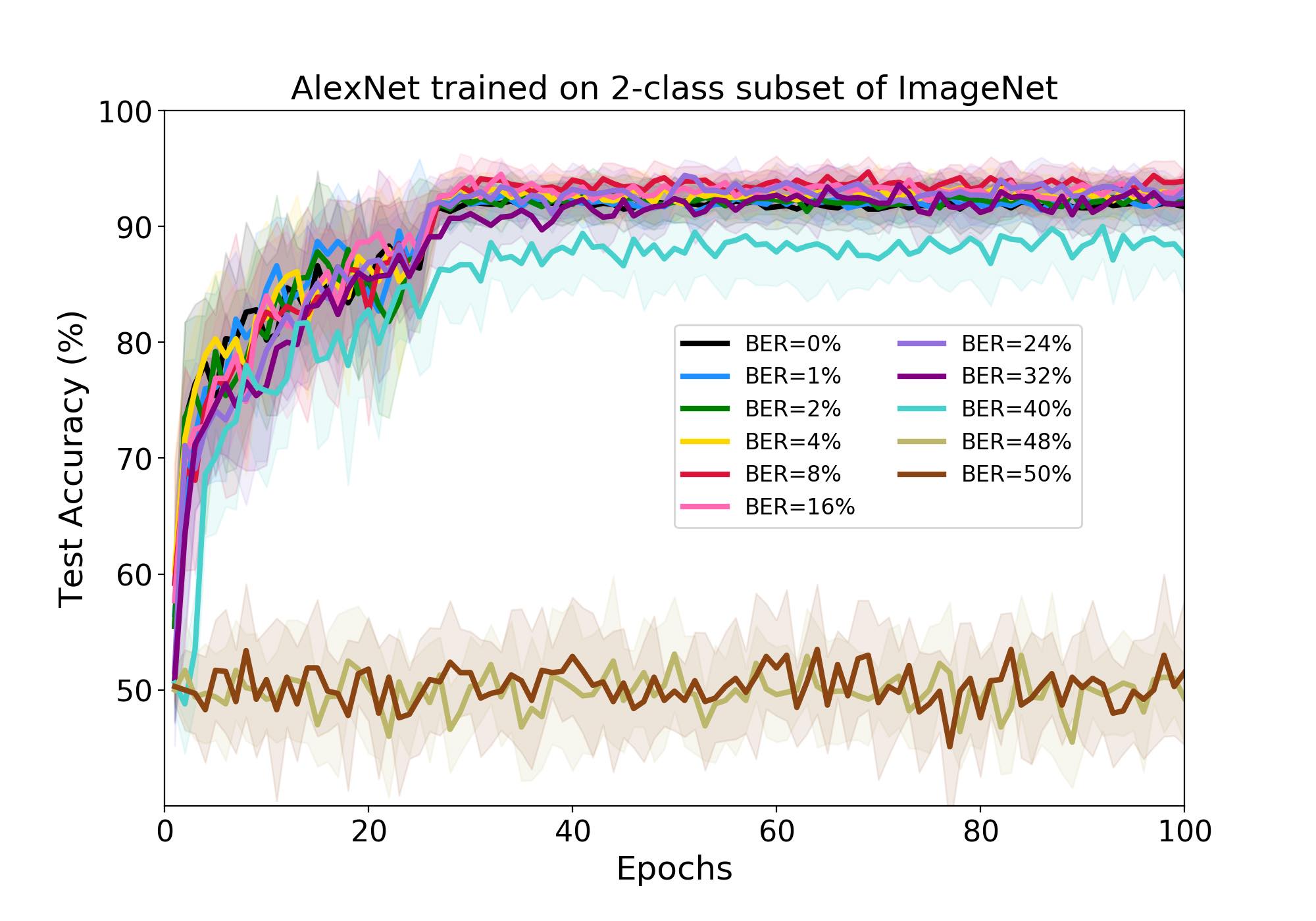}
}

\subfloat[ $\langle \mbox{Top1Max}\rangle_{10}$ for 5 different 2-class subsets of the ImageNet dataset]{\label{subfig:ImageNet_top1max}
 \includegraphics[width=\columnwidth]{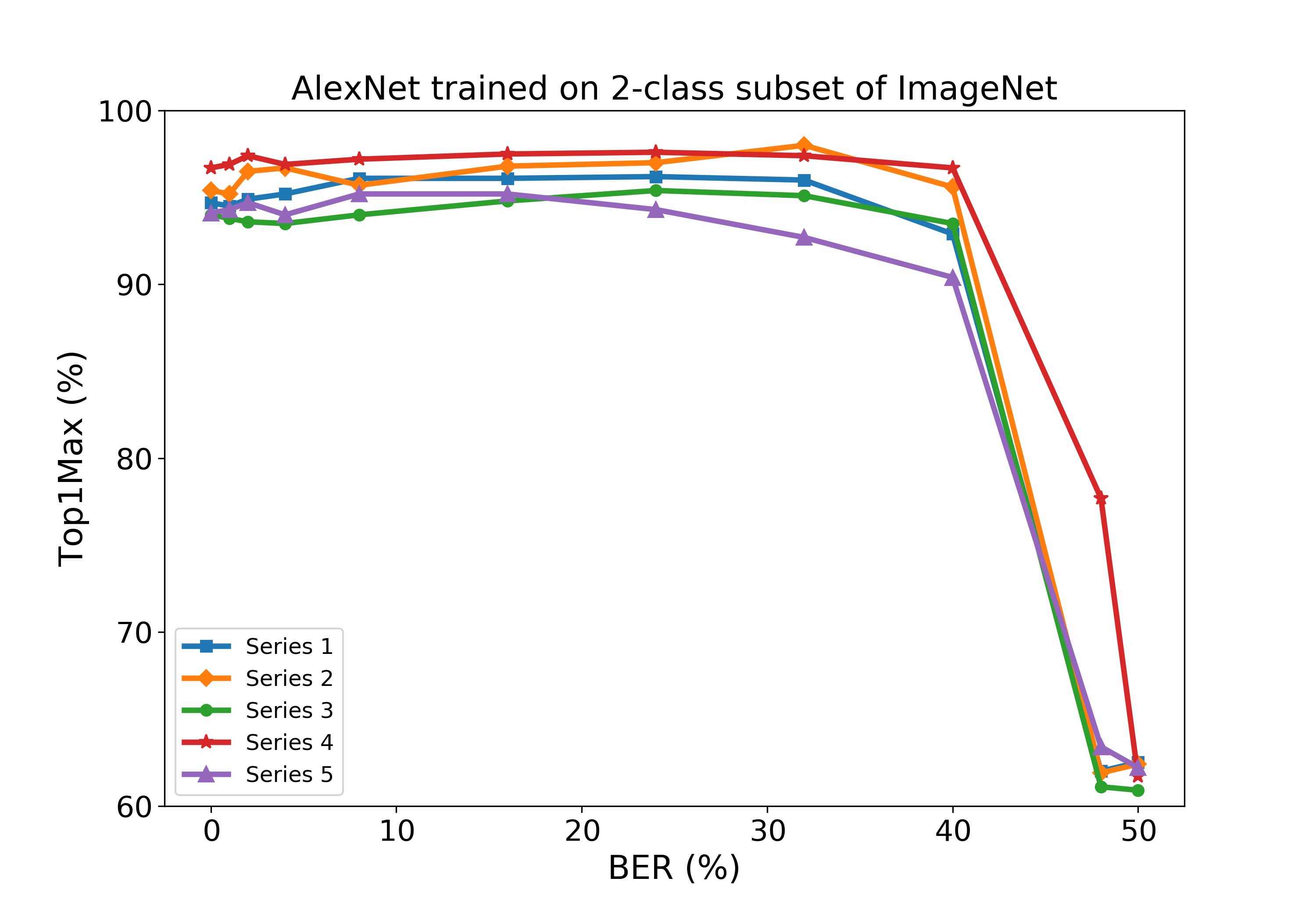}
 }

\caption{ \label{fig:AlexNet2} (a) Binarized AlexNet trained on a 2-class subset of ImageNet with activation errors. We ran 10 experiments and show the average and typical dispersion. (b) Four additional 2-class subsets were studied and the $\langle\mbox{Top1Max}\rangle_{10}$ is shown for each BER value for each of the five cases.}
\end{figure}

Finally we examined the extreme case of a 2-class subset of ImageNet. For each BER value we repeated the experiment 10 times and the mean along with the typical dispersion are shown in Figure \ref{subfig:ImageNet2}. We then randomly selected four additional 2-class subsets and followed the process described above to study the variability of the results. In Figure \ref{subfig:ImageNet_top1max} we show the average Top1Max value for each of the five 2-class subsets and for each BER. Remarkably there was no degradation in test accuracy up to BER of $\sim 32\%$, with $50\%$ being the value that corresponds to complete randomness in the binary activations at which point the test accuracy falls to $\sim 50\%$. 

Overall we observe increasing resilience of the training process to BER in the binary activations as we progressively reduce the number of classes in the system.

\section{Operating MRAM at moderate Write Error Rate}

\subsection{Stochastic errors in MRAM}

In reading or writing an MRAM bit, i.e. an MTJ, there are upper and lower bounds to the voltage amplitude and pulse-length. Specifically: 
\begin{itemize}
\item[$\bullet$]
When reading, the voltage must be high enough and applied long enough to facilitate detection of the MTJ state but not so high/long that it would accidentally switch the MTJ. 
\item[$\bullet$]
When writing, the voltage must be high enough and applied long enough to ensure the information is written correctly but not so high/long that it would excessively stress (or break) the MTJ.
\end{itemize}

In minimizing the error rates we must consider the trade-off between errors and MRAM properties such as speed, density and endurance. For example using long low-amplitude pulses widens the operation windows for both read and write at the cost of speed; increasing the MTJ device diameter makes the device more stable at the cost of lower memory density. The main categories of errors in MRAM bits are the following: 

\begin{enumerate}[label=(\alph*)] 

\item\label{WER_words} \textbf{Write errors}, which occur at a low rate when the voltage amplitude is high and/or the pulse is long enough that the associated spin-polarized current has a high probability of switching the MTJ state. For small devices, where macrospin theory applies, we can determine the switching probability from the voltage pulse and the MTJ parameters using formulas (11)-(12) in Ref. \cite{Tzoufras:2017bx}.

\item\label{BD_words} \textbf{Breakdown} occurs when the voltage amplitude is so high (or the pulse so long) that the MTJ thin insulator material is stressed excessively. Semi-empirical models \cite{Kan:2017rc,Carboni:2018hb} have been developed to describe the device endurance, which is generally found to increase dramatically with the reduction of the voltage amplitude, e.g. using $20\%$ lower write voltage we can raise the number of cycles ($N_c$) by up to $6$ orders of magnitude \cite{Endurance}.

\item \textbf{Retention errors} occur when the MTJ is idle because of spontaneous thermal activation. Small-diameter and/or low-magnetic-anisotropy devices exhibit poor retention. We can calculate the retention error by applying the same formulas as for the write error with zero current. Alternatively we can use the N\'{e}el-Arrhenius model \cite{Neel}.

\item \label{RER_words}  \textbf{Read errors} occur when the voltage amplitude is not high enough (or the read pulse is not long enough) to allow the sense amplifier to detect the resistance state of the MTJ. These errors are not due to the inherent MTJ stochasticity.

\item \label{RDB_words} \textbf{Read disturb errors} occur when the read voltage is so high (or the read pulse so long) that there is a probability of accidentally switching the MTJ while attempting to read it. Read disturb is an inadvertent write and for small devices the read disturb error rate can be calculated with the same formulas as the write error rate.

\end{enumerate}

The operation window for the read process is determined by \ref{RER_words}-\ref{RDB_words} and for writing by \ref{WER_words}-\ref{BD_words}. One of the key advantages of MRAM compared to other nonvolatile memory technologies is its potential to achieve almost unlimited endurance because the number of MTJ write cycles increases rapidly as the ratio $V_{\mbox{write}}/V_{bd}$ reduces, where $V_{bd}$ is the ``breakdown voltage", the value beyond which the MTJ breaks. On the other hand, the Write Error Rate (WER) of the device is a monotonically decreasing function of $V_{\mbox{write}}/V_{c0}$, where $V_{c0}$ is a characteristic ``switching voltage", so that the ratio $V_{\mbox{write}}/V_{c0}$ must be large enough for the WER to attain a specified value. $V_{bd}$ and $V_{c0}$ are both functions of the MTJ parameters.

Special circuit techniques exist to reach $\mbox{WER}\lesssim10^{-15}$  and endurance levels $N_c\gtrsim 10^{13}$, worthy of SRAM and DRAM replacement \cite{TheEngine}. Alternatively, to attain an error rate suitable for applications ($\lesssim 10^{-15}$) the write voltage amplitude must be much higher than $V_{c0}$, or the pulse-length very long ($\tau_{\mbox{write}}\gtrsim 1\mu s$). This is not practical and Error Correction Codes (ECCs) are employed to lower the WER to acceptable levels. Each additional bit of ECC reduces the error rate by 3-4 orders of magnitude but comes at the cost of speed and memory. We can express the \emph{conventional operation window} for the write process in MRAM as:
\begin{equation}\label{eq:ConventionalWindow}
\left.
\begin{array}{c}
\mbox{Low WER}
\\
\mbox{+}
\\
\mbox{High endurance}
\end{array}
\right\}
\Leftrightarrow
[V_{c0} < V_{\mbox{write}} < V_{bd}]
+
\mbox{ECC}
\end{equation}

Yet even with several bits of ECC it can be difficult to achieve sufficiently low WER and high endurance. Instead, we suggest that by operating at moderate WER for certain ANN applications we can dispense with ECC and at the same time reduce $V_{\mbox{write}}$ to raise $N_c$ by many orders of magnitude. We may express the \emph{error-resilient operation window} for the write process in MRAM as:
\begin{equation}\label{eq:ResilientWindow}
\left.
\begin{array}{c}
\mbox{Moderate WER}
\\
+
\\
\mbox{Ultra-high endurance}
\end{array}
\right\}
\Leftrightarrow
V_{c0} \lesssim V_{\mbox{write}}\ll V_{bd}
\end{equation}

Using the low-amplitude voltage values suggested by Eq. (\ref{eq:ResilientWindow}) can boost the endurance, speed and energy efficiency of MRAM.

\subsection{An example of operating at moderate WER}

\begin{figure}[thbp]
\includegraphics[width=9.4cm]{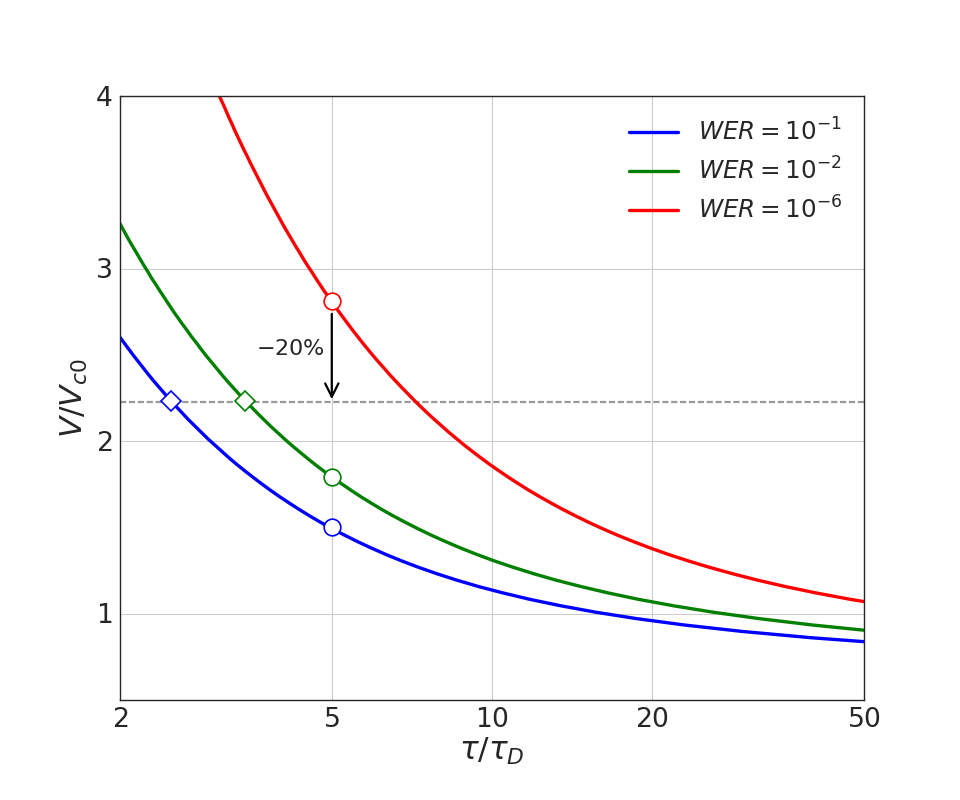}
\caption{\label{fig:Endurance} 
Voltage pulse parameters (amplitude vs pulse-length) for certain WER targets, $0.1$, $0.01$ and $10^{-6}$. Lower WER targets require longer pulses and/or higher amplitude. Relaxing the WER target allows for lower voltage amplitude (circles). Alternatively, at a fixed voltage amplitude we can trade WER for speed (diamonds).} 
\end{figure}

To demonstrate the benefit of operating at moderate WER we present an example using the formulas from Ref. \cite{Tzoufras:2017bx}. We set the normalized energy barrier $\Delta = 40$, i.e. an approximate 1-year retention error of $\exp\Bigl(-[\mbox{(1 year)}/\mbox{(1 ns)}]\exp(-\Delta)\Bigr) \simeq 6 \times10^{-6}$, the characteristic switching time $\tau_D = 2ns$, the switching voltage $V_{c0} = 0.3V$, and the breakdown voltage $V_{bd} = 1.2V$. In Figure \ref{fig:Endurance} we plot the voltage pulse amplitude and duration required for certain WER targets. 

\begin{table}[ht]
\begin{tabular}{|c|cccc|}
\hline
 Target &   $\tau/\tau_D$ & $V/V_{c0}$ & ECC& Endurance ($N_c$) \\
\hline
\hline
 $\mbox{WER} \lesssim 10^{-15}$ &  5& 2.78&2-3 bits&   \\ 
 \hline
 \hline
 $\mbox{WER}=0.01$ &  same  & $-36\%$ & none&unlimited\\ 
$\mbox{WER}=0.1$ &  same  &$-46\%$& none&unlimited\\ 
 \hline
$\mbox{WER}=0.01$ &  $-32\%$ & $-20\%$  &none& $\times10^6$ \\ 
 $\mbox{WER}=0.1$ & $-50\%$ &$-20\%$ & none&$\times10^6$\\
 \hline
\end{tabular}
\caption{\label{tab:Endurance}  The second, third and forth rows correspond to the red, green and blue circles in Figure  \ref{fig:Endurance} and they show the difference in write voltage between error-free memory ($[\mbox{WER}=10^{-6}]+\mbox{ECC}\Rightarrow \mbox{WER}<10^{-15}$) and error-resilient designs at fixed pulse-length. The associated endurance gain in terms of number of cycles is estimated to exceed $10^{10}$.  The bottom two rows show the benefits from error-resilient designs at a $20\%$ reduced voltage (green/blue diamonds in Figure \ref{fig:Endurance}).}
\end{table}

At a fixed pulse-length, relaxing the WER target significantly reduces the write voltage: at $\tau = 5\tau_D = 10ns$ the voltage ($V_{W\!E\!R = 10^{-6}} = 2.78V_{c0}$, red circle) drops by  $36\%$ when the WER target increases from $10^{-6}$ to $10^{-2}$ (green circle) and by $46\%$ when $\mbox{WER} = 0.1$ (blue circle). Such reduction in voltage amplitude enables virtually unlimited number of cycles ($N_c$). Therefore, we can trade back some of the endurance gain for shorter pulse-length. For a constant $20\%$ reduction in voltage, i.e. $0.8V_{W\!E\!R = 10^{-6}} $,  we can calculate the pulse-length required from the $\mbox{WER}  = 0.01$ and $\mbox{WER}  = 0.1$ curves. This yields a $32\%$ and $50\%$ reduction in pulse-length for  $\mbox{WER}  = 0.01$ (green diamond) and $\mbox{WER}  = 0.1$ (blue diamond) respectively,  along with the $20\%$ reduction in voltage amplitude. The comparison against $\mbox{WER} = 10^{-6}$ assumes that a standard MRAM product would employ ECC to lower the WER from $10^{-6}$ down to $10^{-15}$. For the proposed error-resilient operation window no ECC will be used. We summarize these results in Table \ref{tab:Endurance}.

The improvement in energy efficiency when relaxing the WER target can be estimated from the reduction in voltage amplitude and pulse-length seen in Table \ref{tab:Endurance}. At higher speed, i.e. lower $\tau$, the energy savings from relaxing the WER target increase as the WER slopes in Figure \ref{fig:Endurance} become steeper. This is particularly relevant if MRAM is to compete with and complement fast on-chip SRAM.

\section{Conclusions}

Stochasticity is linked in a fundamental and yet not fully understood way to neural networks. At the same time it is an inherent property of MRAM that has hampered it for more than a decade. The  convergence between these two technologies presents a unique opportunity for research and for improving the performance of many ANN applications. 

To demonstrate this we studied the resilience of three binarized ConvNet architectures to errors in the binary activations during the training process. Several image datasets were examined and  the degree of resilience varied significantly across the datasets and the network architectures. For the binarized LeNet-4 and NiN architectures trained on small- and moderate-complexity datasets we found a modest improvement of the error-free test accuracy when the networks were trained with BER of a few percent. The test accuracy gradually dropped when the BER was raised beyond a few percent. For the binarized AlexNet trained on the 1000-class ImageNet dataset we observed a slight degradation in the test accuracy for BER up to $2\%$ followed by a precipitous drop for $\mbox{BER}>4\%$. However, when using subsets of the ImageNet with reduced number of classes, we observed increased error tolerance of the training process. This suggests that the semantic distance between classes is critical in determining the degree of error resilience. The depth and complexity of the network, as well as the number of training images, had no clearly identifiable effect to error resilience. Remarkably, for 2-class subsets of ImageNet, the binarized AlexNet architecture showed no degradation in test accuracy when the network was trained with BER up to $32\%$, with $\mbox{BER} = 50\%$ corresponding to completely random activations.

For MRAM, relaxing the WER targets enables massive improvement in endurance, along with substantially higher speed and energy efficiency. We concentrated the discussion on relaxing the WER because high MRAM endurance is necessary for training. For inference applications we can exploit read, read-disturb and retention errors to improve memory performance---especially for the weights---by increasing memory density and speed.

A more extensive study will include bit errors elsewhere in the system, most notably in the weights, and will allow different error rates for each type of variable. Furthermore, one may use different BER per layer and vary the voltage supply per epoch. A comprehensive understanding of the error resilience of ANNs in different scenarios can open the way for new memory technologies to address many of the pressing hardware challenges. 

\bibliographystyle{plainnat}

\end{document}